\documentclass{nature}

\usepackage{amssymb}
\usepackage{graphicx}
\usepackage{times}
\usepackage{epsfig}
\usepackage{amsmath}
\usepackage{subfigure}
\usepackage{epstopdf}
\usepackage{amssymb}
\usepackage{amsfonts}
\graphicspath{ {pictures/} }
\title{An Adaptive  Framework  for Missing Depth Inference  Using Joint Bilateral Filter}
\author{Rajer Sindhu, Jayesh Ananya}

\begin{document}
\maketitle
\section{ABSTRACT:}
Depth imaging has largely focused on sensor and intrinsics properties. However, the accuracy of acquire pixel is largely dependent on the capture. We propose a new depth estimation and approximation algorithm which takes an arbitrary 3D point cloud as input, with potentially complex geometric structures, and generates automatically a bounding box which is used to clamp the 3D distribution into a valid range. We then infer the desired compact geometric network from complex 3D geometries by using a series of adaptive joint bilateral filters. Our approach leverages these input depth in the construction of a compact descriptive adaptive filter framework. The built system that allows a user to control the result of capture depth map to fit the target geometry. In addition, it is desirable to visualize structurally problematic areas of the depth data in a dynamic environment. To provide this feature, we investigate  a fast update algorithm for the  fragility of each pixel's corresponding 3D point using machine learning. We present  a new for of feature vector analysis and demonstrate the effectiveness in the dataset. In our experiment, we demonstrate the practicality and  benefits of our proposed method by computing accurate solutions captured depth map from different types of sensors and shows better results than existing methods.

\section{Introduction:}
Digital image processing is growing in modern world due to its effectiveness and algorithms which make our day today life very confirmable.
Image sifting for enhancement is a strategy for changing or improving a digital image. Most of the image channels can consider a given image to underscore certain highlights or expel different highlights. Sifting is for the most part including smoothing, honing, and edge improvement. The picture acquired by minimal effort Kinect sensor is having diverse sorts of clamor.
Getting dependable profundity outline fundamental for various applications in picture handling, for example, protest acknowledgment and multi-see rendering. In proposed work versatile directional channels that fill the gaps and smother the clamor inside and out maps. In particular, novel channels whose windows shapes are adaptively balanced in view of the edge heading of the shading picture are exhibited.
Subjective and target investigation comes about demonstrate that our strategy yields higher quality separated profundity maps than other existing strategies, particularly at the edge limits. Keywords: depth map; image filtering; joint bilateral filter; joint trilateral filter; Kinect.

The objective of smoothing a boisterous picture is to stifle clamor, and stress essential highlights. A space-invariant, straight channel performs uniform smoothing and isn't appropriate for safeguarding edges. Since edges contain the primal portray of a picture, it is attractive to protect them. Nonlinear channels that are information versatile were intended to smooth pictures without obscuring the edges.Anisotropic dissemination depicted by Perona and Malik \cite{cite1} was first used to accomplish edge safeguarding smoothing. In this way, Aurich furthermore, Weule utilized nonlinear changes of Gaussian channels \cite{iref2}. Tomasi and Manduchi proposed summed up reciprocal channels whose range channels stifle exceptions to accomplish edge protection \cite{iref3}.

Elad demonstrated that the two-sided channel and anisotropic dispersion develop normally inside a Bayesian structure \cite{iref4}. Various techniques have been proposed for filling the gaps in the profundity outline, are for the most part in view of a pre-adjusted shading picture and in addition the boisterous profundity delineate. In \cite{iref5}, a directional Gaussian channel is utilized to fill the openings by considering the edge data \cite{ref5}.The possibility of directional sifting is promising, however the direction data from edge course in the gap territory isn't sufficiently precise contrasted and the data from edge heading of the comparing shading picture. The gap filling by a settled size window without considering the encompassing locale of the opening might be the reason for the restricted execution.

To use the co-adjusted shading picture and in addition the profundity delineate joint two-sided channel (JBF) \cite{cite6} and the guided channel \cite{cite7} were proposed. In spite of the fact that these strategies can decrease the obscure impact at the edge area \cite{ref10}, the obscuring impact still remains when there is no huge force contrast around profundity discontinuities.

The current approach depicted in \cite{cite8} utilizes an entangled forefront/foundation pixel grouping strategy \cite{cite9} in the fleeting space and applies distinctive JBF channel pieces to the characterized pixels. In spite of the fact that the strategy creates transiently smooth profundity maps, despite everything it experiences the downside and enhances just the forefront objects.The issue of gap filling is additionally considered in \cite{cite10} with the alteration of the notable quick walking based picture inpainting technique \cite{cite11}. Specifically, the shading structure is utilized to decide the weighting capacity for the opening filling \cite{ref15}. The strategies in  \cite{cite12}, be that as it may, create low quality profundity maps if unique profundity and relating shading pictures are not all around adjusted.

An adjusted technique for the joint-trilateral channel (JTF) \cite{cite13} which utilizes both profundity and shading pixels to assess the channel bits is utilized to enhance the nature of both profundity and shading pictures. This technique expect that the profundity outline been prepared to such an extent that there are no opening pixels and the profundity sufficiently delineate quality to be utilized with the shading picture to decide the channel part, which requires an elite calculation for profundity outline handling \cite{ref20}.

The nonlinearity of mark makes the two-sided channel computationally costly in its standard shape. Nonetheless, two-sided channels stay alluring as various works have been devoted to quicken them. Paris and Durand inferred criteria for downsampling in space and force to concoct a quick estimate of the two-sided channel \cite{cite14}. A consistent time calculation for quick reciprocal separating has been proposed. Yang et al. accomplished generous increasing speed at the cost of quantization \cite{cite8}. Adjustments of reciprocal channels have discovered across the board use in various picture handling assignments, for example, denoising \cite{cite9}, brightening remuneration \cite{cite11}, optical-stream estimation, demoaiscking, edge recognition, and so on. Extensive work has additionally been done on enhancing the parameters of the reciprocal channel for enhancing denoising execution. Peng and Rao utilized Stein's unprejudiced hazard gauge (SURE) \cite{cite15} to locate the ideal parameters of the Gaussian respective channel \cite{cite16}. Kishan and Seelamantula accomplished this objective for a two-sided channel with a raised-cosine run piece \cite{cite17}. Chen and Shu utilized Chi-square fair hazard appraise (CURE) \cite{cite17} for improving two-sided channel parameters in squared size MR pictures \cite{cite19}.

 \section{Background}
\subsection{bilateral filtering:}

 Tomasi and Manduchi (1998) proposed non-iterative, neighborhood and straightforward edge-protecting separating technique named as "two-sided sifting". As the name proposes, this sifting was the blend of two sorts of separating, in particular, area sifting and range separating \cite{cite20}.

Area separating authorized geometric closeness between two pixels in light of their spatial region by measuring pixel esteems with coefficients that tumble off with remove. So also, go sifting which implemented photometric likeness between two pixels by averaging picture esteems with weights that rot with uniqueness. Range channels are non-direct in light of the fact that their weights rely upon picture force (for dim scale pictures) or shading (for shading pictures)
Fig.1.

Original image and its Smoothing by Gaussian and Bilateral Filters Choi and Baraniuk (2004) proposed another picture denoising calculation that endeavors a picture's portrayal in different wavelet area. Anticipating a picture onto a Besov chunk of reasonable span relates to a sort of wavelet shrinkage for picture denoising. By characterizing Besov balls in different wavelet spaces and anticipating onto their crossing point utilizing the Projection Onto Convex Sets (POCS) calculation, creators acquired a gauge that adequately joins gauges from various wavelet areas.

Blue and Luisier (2007) proposed an extremely engaging denoising way to deal with picture denoising in view of the standard of SURE. All together for this way to deal with be feasible, the creators included another guideline, that the denoising procedure can be communicated as a direct mix of basic denoising forms—Linear Expansion of Thresholds (LET). The new approach was named as the SURE-LET.

This approach was at first created for monochannel denoising however later been stretched out to shading pictures [Luisier and Blu (2008)], video and blended Poisson-Gaussian commotion condition [Luisier et al. (2011)].

 \section{PROPOSED METHOD :}
\subsection{Adaptive Directional Filters for Depth Image Filtering}
Those Kinect depth sensor experiences two sorts for imperfections: (i) noisy estimations of depth; (ii) holes of unmeasured depth. Our approach with upgrade the blemished profundity picture is with adopt seperate filters for gap and non-hole areas. Likewise demonstrated over figure 2,. The depth image  will be To begin with arranged under hole DD bar h and non-hole DD bar h regions. Then, the filters need aid connected of the non-hole pixels on uproot the depth noise, coming about DD bar h , and then the hole-filling scheme is used to fill the holes, resulting DD bar h . The final depth D bar is the blending of (D bar nh) and (D bar hh).

Since the shading picture I and in addition the profundity delineate is available for the Kinect sensor, the sifting and opening filling misuse the shading picture to find the edge pixels in the profundity outline.

\begin{figure*}[!htb]
\centerline{\epsfig{figure=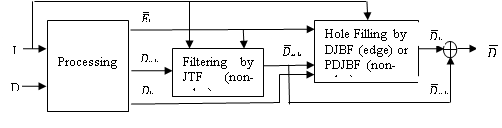, width=15cm}}
\caption{Symmetric constraints by adding four additional edges across left and right limbs}\label{fig:image1}
\end{figure*}

\subsection{5.1	Preprocessing and Edge Detection:}

 Detection Figure 3. Edge detection and hole expansion in the preprocessing block of Figure 1.
 \begin{figure*}[!htb]
\centerline{\epsfig{figure=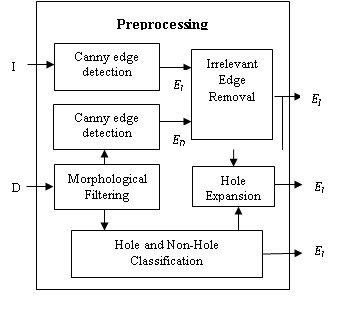, width=7cm}}
\caption{Symmetric constraints by adding four additional edges across left and right limbs}\label{fig:image1}
\end{figure*}
Likewise indicated in Figure 3, the input depth map D is first pre-processed to remove small depth hole pixels that show up haphazardly between sequential frames. To this end, morphological closing operation with 5 × 5 mask is applied to D, yielding the outlier-removed depth map. For the simplicity, let D hereafter denote the pre-processed depth map.

\subsection{5.2 Filtering Non-Hole Regions :}

Since the clamor and the gaps in the profundity picture are dealt with independently, we have to group the profundity picture into non-opening (Dnh) and gap (Dh) zones Note that the Kinect sensor gives labels from guaranteeing non-available to the pixels for no returned flags Furthermore we organize these non-available pixels Likewise gap pixels. 
\begin{figure*}[!htb]
\centerline{\epsfig{figure=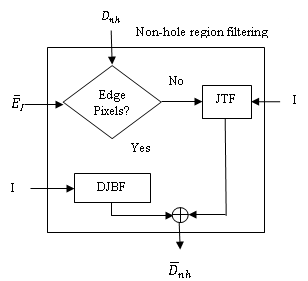, width=10cm}}
\caption{Symmetric constraints by adding four additional edges across left and right limbs}\label{fig:image1}
\end{figure*}
Figure 3 indicates the flowchart of the non-hole region filtering. Figure 4. The flowchart of the non-hole region filtering.
The non-hole depth region h is usually noisy and the conventional JBF [6] can be considered for the noise filtering, which processes a pixel at position using a set of neighbor pixels in the window

\[ \overline{D_{nh}} = \]
$$\sum_{q\epsilon\phi^p} D_{nh}(q)f_{s}^{d}(q_{x}-p_{x},q_{x}-p_{y})f_{r}^{c}(I_{}^{p}-I_{}^{q})   (1)$$
\[f_{s}^{d}(q_{x}-p_{x},q_{x}-p_{y}) = e^-\frac{1}{2}(\frac{(q_{x}-p_{x})^2+(q_{x}-p_{y})^2}{\sigma_{s}^{2}})  (2)
\]
\[ f_{r}^{c}(I_{}^{p}-I_{}^{q}) = e^-\frac{1}{2}(\frac{I_{}^{p}+I_{}^{q}}{\sigma_{r}})^2  (3)
\]
In particular, for the non-edge region in  E1 
\[ \overline{D_{nh}} = \]
$$\sum_{q\epsilon\phi^p} D_{nh}(q)f_{s}^{d}(q_{x}-p_{x},q_{x}-p_{y})f_{r}^{c}(I_{}^{p}-I_{}^{q})f_{r}^{d}(D_{nh}(p)-D_{nh}(q))   (4)$$
In equastion (4) we consider the depth similarity around the neighborhood pixels as:
\[ f_{r}^{d}(D_{nh}(p)-D_{nh}(q)) = e^-\frac{1}{2}(\frac{D_{nh}(p)-D_{nh}(q)}{\sigma_{r}})^2  (5) \]
we have the DJBF as:
\[ \overline{D_{nh}} = \]
$$\sum_{q\epsilon\phi^p} D_{nh}(q)f_{ds}^{d}(q_{x}-p_{x},q_{x}-p_{y})f_{r}^{c}(I_{}^{p}-I_{}^{q})   (6)$$
where the directional Gausian filter (DGF) is used for the spatial filter kernel as follows:
\[ f_{ds}^{d}(q_{x}-p_{x},q_{x}-p_{y}) = e^-\frac{1}{2}(\frac{x_{\theta}^{2}}{\sigma_{x}^{2}}+\frac{y_{\theta}^{2}}{\sigma_{y}^{2}})  (7) \]
\[x_{\theta} =(q_{x}-p_{x})\cos\theta-(q_{y}-p_{y})\sin\theta \]
\[y_{\theta} =(q_{x}-p_{x})\sin\theta-(q_{y}-p_{y})\cos\theta \]

The edge direction theta is given by:
\[\theta = \tan^{-1}(g_{x}/g_{y})   (8) \]
\subsection{5.3 Hole Filling:}

In the wake of sifting the profundity pixels to acquire more sure profundity esteems in the non-gap districts, those separated profundity information are utilized to fill the openings. In the first place, to decide the source of the gaps, we misuse the edge data E I again to characterize the openings into edge or non-edge areas.
\begin{figure*}[!htb]
\centerline{\epsfig{figure=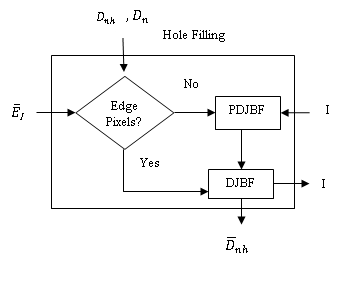, width=10cm}}
\caption{Symmetric constraints by adding four additional edges across left and right limbs}\label{fig:image1}
\end{figure*}
For the openings in the non-edge locale we propose a fractional directional joint reciprocal channel (PDJBF) to fill the gap pixels, though the DJBF in Equation (6) is utilized to fill the gap pixels in the edge area. To upgrade the execution of the gap filling at the edge locales, the opening pixels in the non-edge area are filled first.

Figure 4 demonstrates the flowchart of the proposed opening filling strategy. Figure 5. The flowchart of the hole region filling. Once the heading of the opening filling is resolved, to fill the gap pixels in the non-edge area, the proposed PDJBF utilizes the DGF as a spatial bit, which can smooth pictures while holding the edge points of interest
The PDJBF is characterized as takes after:

\section{CONCLUSION:}

The new system which is abusing the edge data for the profundity delineate is proposed in this approach effectively. We permit the nearby help of the window for the separating to shift adaptively as indicated by the heading of the edge and the relative position between the edge removed from the shading picture and the to-be-sifted pixel too. Our sifting approach is actualized for the opening filling issue of the Kinect profundity pictures.

The proposed technique demonstrated that utilizing the versatile directional channel portion with versatile channel run gives better opening filling comes about particularly for the gap pixels close protest limits. The viability of the proposed technique was shown quantitatively by utilizing the manufactured test pictures and subjectively utilizing the Kinect test pictures.

In this way, at last subjective and additionally target quality outcomes will demonstrate that proposed work is powerful for profundity outline.


 \end{document}